\documentclass[journal]{IEEEtran}

\usepackage{graphicx}
\usepackage{amsmath}
\usepackage{amsfonts}
\usepackage{mathtools}
\usepackage{amsmath}
\usepackage{diagbox}
\usepackage{multirow}
\usepackage{float}

\usepackage{color, soul}
\usepackage{icomma}
\usepackage[range-phrase=--]{siunitx}
\usepackage{algorithm, algpseudocode}
\usepackage{tabularx}
\usepackage{booktabs}

\usepackage{xargs}    
\usepackage[pdftex,dvipsnames]{xcolor} 
\usepackage[colorinlistoftodos,prependcaption,textsize=tiny]{todonotes}
\usepackage{multicol}

\newcommandx{\improvement}[2][1=]{\todo[linecolor=Plum,backgroundcolor=Plum!25,bordercolor=Plum,#1]{#2}}

\usepackage{glossaries}
\newacronym{MLC}{MLC}{multi-label scene classification}
\newacronym{SLC}{SLC}{single-label classification}
\newacronym{RS}{RS}{remote sensing}
\newacronym{CV}{CV}{computer vision}
\newacronym{LP}{LP}{label propagation}
\newacronym{xAI}{xAI}{explainable artificial intelligence}
\newacronym{GAN}{GAN}{generative adversarial network}
\newacronym{SAR}{SAR}{synthetic aperture radar}
\newacronym{CAMs}{CAMs}{class activation maps}
\usepackage{booktabs}

\makeatletter
\newcount\SOUL@minus 
\makeatother

\usepackage{etoolbox}
\usepackage{overpic}

\def\subcaptionbelow{subcaptionbelow}

\def\subcaptionstyle{subcaptionbelow}

\usepackage{cite}
\usepackage{hyperref}
\usepackage[
    singlelinecheck=false, 
    font=small,
]{caption}
\usepackage[slc=off, subrefformat=parens]{subcaption}

\ifx\subcaptionstyle\subcaptionbelow
\else
\fi

\usepackage[capitalise,noabbrev]{cleveref}

\begin{document}

\title{Self-Supervised Cross-Modal Text-Image Time Series Retrieval in Remote Sensing}

\author{Genc~Hoxha,~\IEEEmembership{~Member,~IEEE,} Olivér Angyal
        and~Begüm~Demir,~\IEEEmembership{Senior~Member,~IEEE}
\thanks{Genc Hoxha and Beg{\"u}m Demir are with the Faculty of Electrical Engineering and Computer Science, Technische Universit{\"a}t Berlin, 10623 Berlin, Germany and also with the Berlin Institute for the Foundations of Learning and Data (BIFOLD), 10623 Berlin, Germany (emails: genc.hoxha@tu-berlin.de, demir@tu-berlin.de).

Olivér Angyal is with the Faculty of Electrical Engineering and Computer Science, Technische Universit{\"a}t Berlin, 10623 Berlin, Germany (e-mail: \mbox{oliver.angyal@gmail.com}).}

}

\markboth{Journal of \LaTeX\ Class Files,~Vol.~14, No.~8, August~2015}%
{Shell \MakeLowercase{\textit{et al.}}: Bare Demo of IEEEtran.cls for IEEE Journals}

\maketitle

\begin{abstract}
The development of image time series retrieval (ITSR) methods is a growing research interest in remote sensing (RS). Given a user-defined image time series (i.e., the query time series), the ITSR methods search and retrieve from large archives the image time series that have similar content to the query time series. Existing ITSR methods in RS are designed for unimodal retrieval problems, 
relying on an assumption that users always have access to a query image time series in the considered image modality. In operational scenarios, this assumption may not hold. To overcome this issue, as a first time in RS we introduce the task of cross-modal text-image time series retrieval (text-ITSR). In particular, we present a self-supervised cross-modal text-ITSR method that enables the retrieval of image time series using text sentences as queries, and vice versa. In detail, we focus our attention on text-ITSR in pairs of images (i.e., bitemporal images). Our text-ITSR method consists of two key components: 1)  modality-specific encoders to model the semantic content of bitemporal images and text sentences with discriminative features; and 2) modality-specific projection heads to align textual and image representations in a shared embedding space. To effectively model the temporal information within the bitemporal images, we exploit two fusion strategies: i) global feature fusion (GFF) strategy that combines global image features through simple yet effective operators; and ii) transformer-based feature fusion (TFF) strategy that leverages transformers for fine-grained temporal integration. Extensive experiments conducted on two benchmark RS archives demonstrate the effectiveness of our method in accurately retrieving semantically relevant bitemporal images (or text sentences) to a query text sentence (or bitemporal image). The code of this work is publicly available at https://git.tu-berlin.de/rsim/cross-modal-text-tsir. 
 \end{abstract}


\IEEEpeerreviewmaketitle

\section{Introduction}\label{sec:introduction}
\IEEEPARstart{T}{\lowercase{he}} rapid advancements in remote sensing (RS) technology have resulted in an unprecedented growth of image archives that usually contain multitemporal images, i.e., images acquired over the same geographical area at different times. Accordingly, the development of content-based image time series retrieval (ITSR) methods that aim to query specific kinds of change relevant to the user from such archives has attracted great attention in RS. ITSR can be divided into two different categories: 1) retrieval of long-term changes (e.g., seasonal changes or time varying phenomena that can be observed at different time resolution); and 2) retrieval of short-term changes (e.g., forest fires, floods) \cite{Bovolo_TSR_15}.

Any ITSR method essentially consists of (at least) two steps: 1) description of each image time series by a set of features; and 2) retrieval of time series of images similar to the query time series. Querying image time series from large RS data archives depends on the capability and effectiveness of the techniques in describing and representing the images \cite{Bovolo_TSR_15,Vuran_deep_22}. In the RS literature, several methods have been presented for ITSR purposes. As an example, Bovolo et al.\cite{Bovolo_TSR_15} propose to exploit spectral change vector analysis to model the semantic content of bitemporal images. Subsequently, they utilize the k-means clustering algorithm and Euclidean distance to perform bitemporal image retrieval, specifically for the retrieval of short-term changes. Ma et al. \cite{ma2017content}, on the other hand, employ several hand-crafted color and texture features to represent bitemporal images. However, hand-crafted features are not capable of representing the rich semantic information contained in the image time series, resulting in limited retrieval performance. To address this issue, deep neural networks (DNNs) have shown their effectiveness in learning discriminative image representations directly from raw images, becoming the default selection for image representation learning in many RS applications \cite{DLRS,TuiaAI}. As an example, Vuran et al. \cite{Vuran_deep_22} investigate two methods based on DNNs to learn bitemporal image representations. The first method is based on deep change vector analysis \cite{saha_19_DCVA} and consists of exploiting feature differences from different layers of convolutional neural networks (CNNs). The second method is based on autoencoders and aims to reconstruct the difference image from the bitemporal images where the latent space of the autoencoder is used as bitemporal image representations. Retrieval is then performed by measuring the similarity  between the features of the query bitemporal images and those of the archive images \cite{Vuran_deep_22}. 

The above-mentioned methods are defined for single-modality ITSR problems (i.e., uni-modal ITSR). For a given query image time series, uni-modal retrieval systems search for the image time series with semantically similar contents from the same modality image archive. 
However, multi-modal data archives, including different modalities of satellite image time series as well as textual data, are currently available. Thus, the development of retrieval systems that return a set of semantically relevant results of different modalities given a query in any modality (e.g., using a text sentence to search for image time series) has recently attracted great attention in RS. 

In this paper, we introduce in RS the cross-modal text-image time series retrieval (text-ITSR) task, where queries from one modality (e.g., text) can be matched to archive entries from another (e.g., image time series in RS).  Although text image retrieval is widely studied for single-date images (see Section \ref{related_work}), to the best of our knowledge, no prior work explored its extension to time series of images in RS. The task of retrieving image time series based on natural language queries plays a critical role in effectively handling large-scale RS archives for searching long-term as well as short-term changes. Using text sentences as queries allows users to express complex spatial-temporal semantic concepts without requiring an exact image time series example. This significantly broadens the usability of ITSR systems in operational scenarios, making them more intuitive and adaptable to user needs.Examples of such scenarios include environmental monitoring, disaster response, or urban development. In environmental monitoring, users can search for deforestation areas by typing a query text as 'the forest has been gradually cleared,' without any need for an exact image time series example. Similarly, queries like 'the areas are flooded' can be used to search for flooded regions, and 'new buildings have been constructed' to identify urban expansion."
We would like to note that text-ITSR is inherently more challenging than conventional text-image retrieval, since it requires modeling the content of the image time series and to establish proper associations with the linguistic information presented in the temporal order. In this paper, to address the text-ITSR problem, we present a self-supervised cross-modal text-ITSR method. Specifically, we devote our attention in the retrieval of short term changes and thus we consider pairs of bitemporal images and their text sentences. The presented text-ITSR method consists of two main modules: 1) modality-specific encoders (which extract discriminative features from both text and bitemporal images; and 2)  modality-specific projection heads (which learn a joint feature representation space between the bitemporal images and text sentences by employing the contrastive loss) (see Fig. {\ref{Fig:scheme}}, ). To effectively model the semantic content of the bitemporal images, we exploit two main fusion strategies: i) global feature fusion (GFF) strategy that combines the global features extracted from bitemporal images using two simple but effective operators that are feature concatenation or element-wise feature subtraction; and ii) transformer-based feature fusion (TFF) that leverages the transformer architecture to combine the bitemporal information. Extensive experiments carried out on two RS benchmark archives composed of pairs of text and bitemporal images demonstrate the effectiveness of our method for cross-modal text-bitemporal image retrieval. We further compare our method with a baseline based on an early fusion strategy, which stacks the bitemporal images before feature extraction and applies contrastive learning for cross-modal alignment. The results show that our method outperforms the baseline, confirming its effectiveness for text-bitemporal image retrieval.

The remaining part of this paper is organized as follows. Section \ref{related_work} presents the related work on (single-date) text-image retrieval in RS. Section \ref{methods} introduces the presented method. Section \ref{dataset_setup} describes the considered RS image archives and the experimental setup, while the experimental results are presented in Section \ref{experimental_results}. Finally, in Section \ref{conclusion}, the conclusion of the work is drawn. 
\section{Related Work} \label{related_work}

Our paper presents the first study in RS in the context of text-ITSR, whereas text-image retrieval (achieved on analyzing single-date images without considering temporal content) is widely investigated in RS. In detail, text-image retrieval aims at retrieving semantically relevant single-date RS images to a given user-defined query text, and vice versa \cite{Hoxha_2020,Abdullah_2020,rahhal2020deep,SAM_21,mikriukov2022deep,Rahhal_Transf_22,mikriukov2022unsup,mi2024knowledge,promt_txt_img}. Early works utilize image captioning systems to generate text sentences for RS images and employ text matching techniques to perform the retrieval \cite{Hoxha_2020}. As an example, Hoxha et al. \cite{Hoxha_2020} propose a retrieval system that generates and exploits text sentences for text-image retrieval. Their system utilizes an image captioning system as a combination of CNNs and long short-term memory (LSTM) \cite{LSTM_97} to generate text sentences of RS images, enabling image-text retrieval based on text similarity. The resulting system allows users to perform image and text retrieval by utilizing either an image as a query (for which a text sentence is generated) or directly textual queries. However, this two-step approach is highly dependent on the ability of the image captioning system to generate accurate captions of the query and archive images.

Recent works focus primarily on learning a joint image and text representation space, where images and their associated text are projected close to each other \cite{Abdullah_2020,rahhal2020deep,SAM_21,mikriukov2022deep,Rahhal_Transf_22,mikriukov2022unsup,mi2024knowledge}. 
In these works, modality-specific encoders are first used to represent images and text with discriminative features. These features are then further optimized and projected into the learned joint embedding space. The joint embedding space is learned through contrastive objectives such as pairwise contrastive loss \cite{pairwise_contrast_06} or triplet loss \cite{tripletLoss_15}. For example, Abdullah et al. \cite{Abdullah_2020} propose a deep bidirectional triplet network that consists of a CNN encoder for images and an LSTM encoder for text. Then the triplet loss is used to learn a common embedding space, ensuring that semantically similar images and texts are projected close to each other, while dissimilar ones are placed farther apart. As in \cite{Abdullah_2020}, CNNs and LSTMs are employed in \cite{rahhal2020deep} as image and text encoders whereas the shared embedding space is learned via an unsupervised training strategy using pairwise contrastive loss. A semantic alignment module is proposed in \cite{SAM_21} that combines attention and gating techniques for a more representative joint embedding space. The attention mechanism is used to find and align image and textual correspondences, while the gating mechanism filters unnecessary information. Yuan et al.\cite{Yuan_multi_scaletriplet} present an asymmetric multi-modal matching network that consists of a multi-scale visual self-attention module and a dynamic filtering function. The self-attention module captures multi-scale feature representations from RS images, while the dynamic filtering function removes redundant features. The authors later refine their approach and present a lightweight multi-scale method optimized via knowledge distillation and contrastive loss enhancing the retrieval speed  \cite{Yuan_multi_distilaton}. In addition, Yuan et al. \cite{Yuan_globalLocal22_triplet} present a global-local information fusion module, which combines local features extracted via a graph convolutional neural network (GCNN) and global features from a CNN to generate multi-scale visual representations. 

The development of Transformer-based models \cite{vaswani2017attention}, especially those leveraging the CLIP foundation model \cite{Radford2021LearningTV}, have gained growing attention for visual-language alignment in RS \cite{Rahhal_Transf_22,YUAN_Transf_block23,promt_Transf_2024,hu_GlobaL_LOcal_transf24,Zhao_Transf_24_MAsked,Pyramid_transformers,eventretrieval_Tansf}. As an example, Rahhal et al. \cite{Rahhal_Transf_22} propose a multi-language text-image retrieval approach that consists of pre-trained language and vision transformers (i.e., CLIP \cite{Radford2021LearningTV})  to extract text and image representations. Their approach supports retrieval in multiple languages extending beyond English. Yuan et al. \cite{YUAN_Transf_block23} propose a parameter-efficient transfer learning approach based on CLIP to effectively and efficiently transfer visual–language knowledge from the natural domain to the RS domain for text–image retrieval.  Hu et al. \cite{hu_GlobaL_LOcal_transf24} propose a CLIP-based soft-alignment approach to align global and local image and text features utilizing pairwise contrastive loss. Zhao et al. \cite{Zhao_Transf_24_MAsked} introduce a masked interaction inferring and aligning (MIIA) module \cite{Zhao_Transf_24_MAsked} that is integrated into CLIP to learn fine-grained image-text features. The MIIA module is defined as a combination of self- and cross-attention modules to establish connections between image and text representations by predicting masked image and text tokens.

Although the above-mentioned methods are effective for text-image retrieval in RS, they are not suitable for text-ITSR due to their inability to characterize the temporal content and to exploit it for retrieval problems.

\section{Proposed  Method}\label{methods}
In this paper, we introduce cross-modal text-ITSR task. In particular, we focus our attention on the retrieval of the abrupt changes. Let $\mathcal{D} =\{(\boldsymbol{X}^i,\boldsymbol{Y}^i) \}^N_{i=1}$ denote an unlabeled multi-modal training set consisting of $N$ bitemporal images and text pairs. Let $\boldsymbol{X}^i = (\boldsymbol{I}_{t_{1}}^i,\boldsymbol{I}_{t_{2}}^i)$ be the $i^{th}$ pair of co-registered RS images acquired over the same geographical area at times $t_1$ and $t_2$, respectively. Let $\boldsymbol{Y}^i = (w_1, w_2,...,w_M)^i$ be the corresponding text sentence (with $M$ ordered words $w$) describing the changes between the co-registered images in $\boldsymbol{X}^i$. We assume that a labeled training set is not existing.  To achieve text-ITSR, we present a self-supervised method for cross-modal retrieval, where queries from one modality (e.g., text) can be matched to archive entries from another (e.g., image time series).  To this end, we first extract discriminative features from bitemporal images and their corresponding textual descriptions utilizing modality-specific encoders. For the text modality, we adopt the pretrained Bidirectional Encoder Representations from Transformers (BERT) \cite{BERT} as text encoder. Given an input sentence $\boldsymbol{Y}^i$, we obtain its feature representation as:
\begin{equation}
    \label{eq:in_txt}
    f^{\boldsymbol{Y}^i} =  BERT(\boldsymbol{Y}^i).
\end{equation}
For the image modality, we utilize the Vision Transformer (ViT)\cite{ViT} to extract features from each image $\boldsymbol{I}_{t_{1}}^i,\boldsymbol{I}_{t_{2}}^i$ in the bitemporal pair $\boldsymbol{X}^i$. Each image is divided into non-overlapping patches, which are then processed by a transformer encoder: 
\begin{equation}
    \label{eq:vit_in_t1}
    f^{\boldsymbol{{I}_{t_{1}}^i}}= ViT (\boldsymbol{I_{t_{1}}^i})
\end{equation}

\begin{equation}
    \label{eq:vit_in_2}
    f^{\boldsymbol{{I}}_{t_{2}}^i}= ViT (\boldsymbol{{I}}_{t_{2}}^i)
\end{equation} 

\begin{figure}
    \centering
    \includegraphics[scale =0.35]{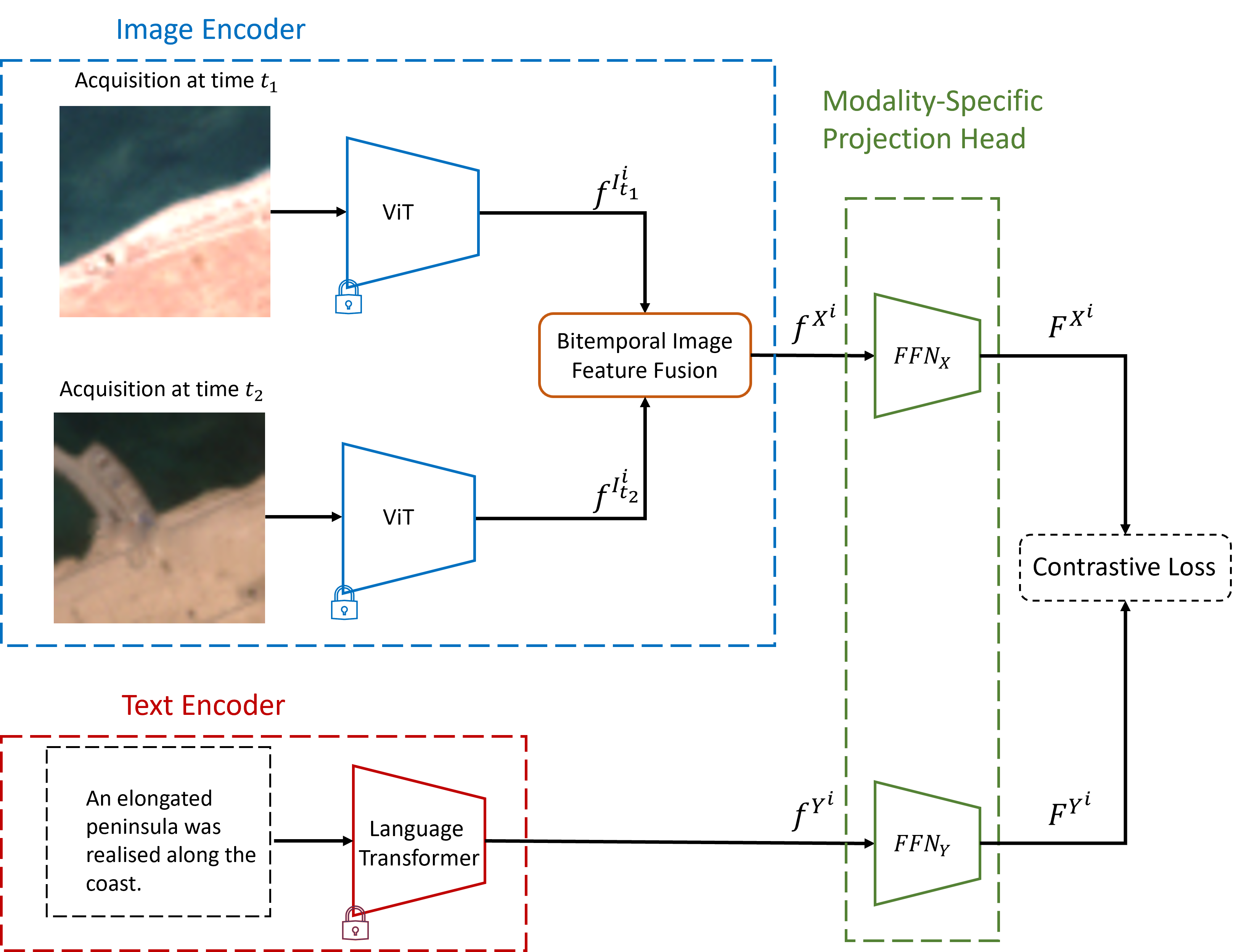}
    \caption{Overall architecture of the proposed text-ITSR method.}
    \label{Fig:scheme}
\end{figure}

To model the  semantic changes between the two image acquisitions, we exploit two fusion strategies that are applied at the feature level (see Fig. {\ref{Fig:scheme}}): i) Global Feature Fusion (GFF) which combines the global  features extracted independently from each image in the bitemporal pair either by concatenation, preserving complementary temporal information, or by element-wise subtraction, emphasizing semantic changes \cite{FC_Conc_Differ_2018,survey_CD_25}.
ii) Transformer-based Feature Fusion (TFF) which leverages a cross-attention mechanism on patch-level features extracted independently from each image in the bitemporal pair \cite{LiuCCD}. This allows to model spatial and spectral information independently within each image. Then, to model the temporal information between the multitemporal images, we apply the cross-attention mechanism that enables the patches from one timestamp to attend to those from the other. This facilitates interaction between spatial regions over time and enhances the model's ability to identify fine-grained semantic changes.

After applying one of the fusion strategies to the bitemporal image features, the resulting image representations and the previously extracted textual features are projected via modality-specific heads into a shared embedding space.
This space is learned using a contrastive learning objective that encourages semantically aligned image-text pairs to be projected close to each other, while pushing apart those that are not semantically related.  Through this objective, the model aligns the temporally structured language with the temporal content of bitemporal imagery, enabling a robust joint representation for cross-modal retrieval. In the following we provide a detailed description of our fusion strategies as well as the optimization of our text-ITSR method.


\subsection{Global Feature Fusion (GFF)}
To achieve global feature fusion (GFF) we exploit the global features obtained by ViT. To this end, we present two simple but effective operators for GFF: 1) element-wise feature subtraction denoted as GFF: Subtraction and 2) feature concatenation denoted as GFF: Concatenation. In detail, the GFF: Subtraction can be expressed as follows:
\begin{equation}
    \label{eq:sub_GFF}
    f^{\boldsymbol{X}^i} =  f^{\boldsymbol{{I}}_{t_{2}}^i}- f^{\boldsymbol{{I}}_{t_{1}}^i}
\end{equation}
whereas the GFF: Concatenation is defined as:
\begin{equation}
    \label{eq:conc_GFF}
    f^{\boldsymbol{X}^i} =  Concat(f^{\boldsymbol{{I}}_{t_{2}}^i}, f^{\boldsymbol{{I}}_{t_{1}}^i}).
\end{equation}
These fusion strategies are parameter-free and computationally efficient, offering a strong and interpretable baseline for bitemporal image representation learning for text-ITSR retrieval.

\subsection{Transformer-based Feature Fusion (TFF)} 
\begin{figure}
    \centering
    \includegraphics[scale =0.36]{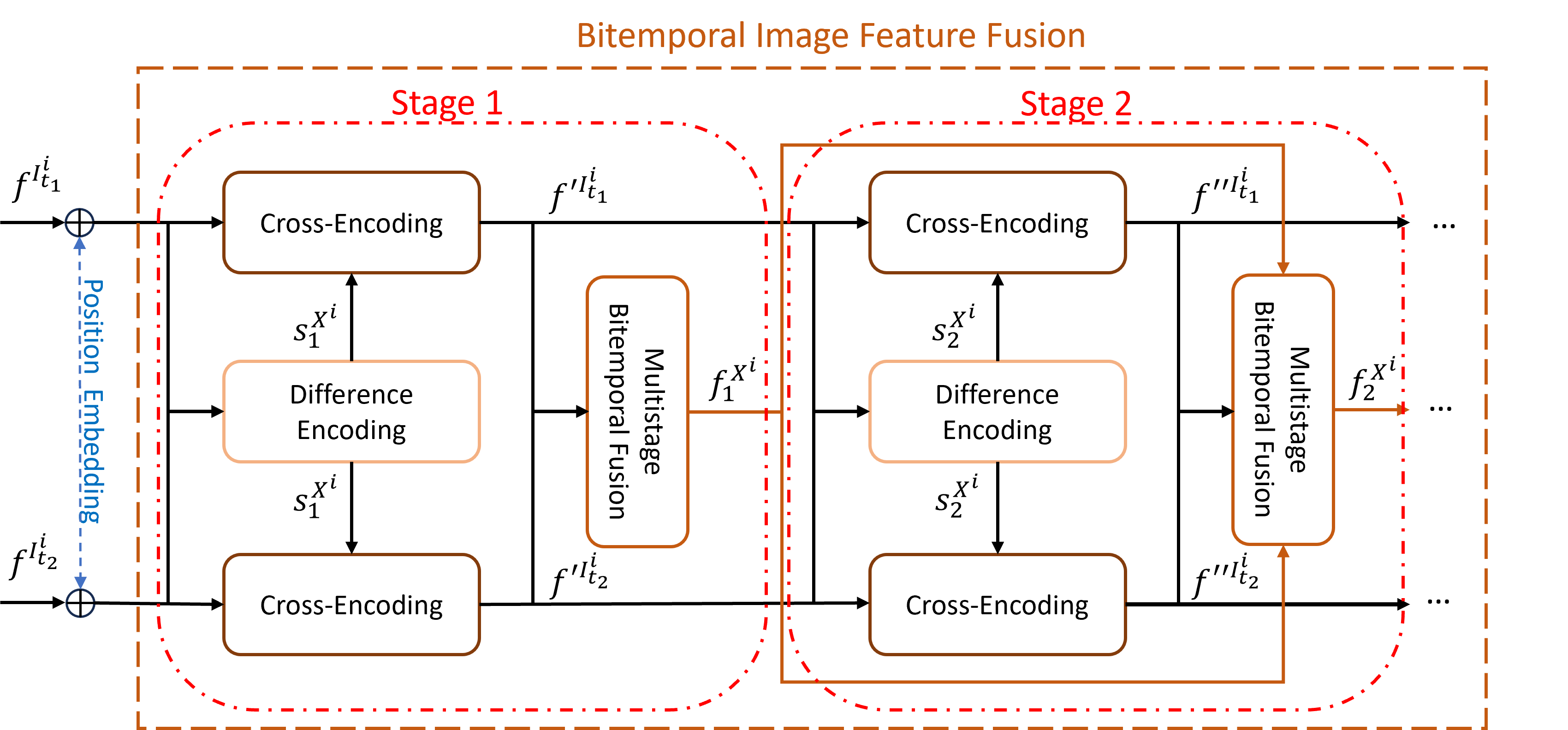}
    \caption{Architecture of the feature fusion of bitemporal images.}
    \label{Fig:bi_fusion}
\end{figure}

While GFF focuses on global-level representations using the ViT class tokens, the Transformer-based Feature Fusion (TFF) strategy operates on spatially localized patch embeddings, enabling a more detailed modeling of temporal changes. Accordingly, it processes and fuses through multiple stages the patch embedding $f^{\boldsymbol{{I}}_{t_{2}}^i}, f^{\boldsymbol{{I}}_{t_{1}}^i} \in \mathbb{R}^{T\times d_{E_I}}$  obtained from the bitemporal images. Fig. \ref{Fig:bi_fusion} shows an illustration of the TFF strategy that is mainly based on \cite{LiuCCD} with some adaptation to the ViT image encoder. To capture temporal differences between bitemporal images, TFF begins by computing the element-wise difference between their patch embeddings
\begin{align}
    \label{eq:sub_TFF}
    s^{\boldsymbol{X}^i} =  f^{\boldsymbol{{I}}_{t_{2}}^i}- f^{\boldsymbol{{I}}_{t_{1}}^i}  && s^{\boldsymbol{X}^i} \in \mathbb{R}^{T\times d_{E_I}}
\end{align}
where $T$ and $d_{E_I}$ are the total number of patches and the patch embedding dimensions, respectively. To extract correlations between the patch embeddings, the difference of the patch embedding $s^{\boldsymbol{X}^i}$  is passed through a cross-attention layer to extract correlation across patches:
\begin{equation}
    \label{eq:cross_TFF}
    CrossAttention(Q,K,V) = softmax(\frac{QK^T}{\sqrt{d}})V
\end{equation}
where the image pair patch embeddings $f^{\boldsymbol{{I}}_{t_{1}}^i},f^{\boldsymbol{{I}}_{t_{2}}^i} \in \mathbb{R}^{T\times d_{E_I}}$ serve as query $Q$, their difference $s^{\boldsymbol{X}^i} \in \mathbb{R}^{T\times d_{E_I}}$ vector as key $K$ and value $V$.  With the scaled dot product of $Q$ and $K$, an attention score is calculated between each pair of vectors in the input sequence. Intuitively, this is a measure of similarity, that is, how relevant the query vector is to the key vector. For multitemporal images, since the input sequence includes patch embeddings from the two image acquisitions (i.e., bitemporal images), these attention scores capture the temporal correlation between patches. The scaling by $\sqrt{d}$ ($d$ being the internal hidden dimension) is a normalization factor that tends to make training more stable \cite{vaswani2017attention}. Through the softmax function, normalized attention scores are obtained for each vector. Finally, multiplying the normalized attention scores with $V$ produces the context-aware representation of each vector. 

As in the standard transformer architecture \cite{vaswani2017attention}, the cross-attention mechanism  defined in (\ref{eq:cross_TFF}) is applied across $n$ parallel attention heads. Each head has its own set of learnable projection matrices for queries, keys, and values, enabling the model to attend to different aspects of the input representations. The outputs of all attention heads are concatenated and multiplied with a separate learnable weight matrix $W^O \in \mathbb{R}^{nd\times d_{E_I}}$ in order to match the input dimensions of the next layer as described in  (\ref{eq:cross_TFF_multihead}) and (\ref{eq:cross_TFF_cross_multi})
\begin{equation}
    \label{eq:cross_TFF_multihead}
    MultiHead(f^{\boldsymbol{{I}}_{t_{2}}^i}, f^{\boldsymbol{{I}}_{t_{1}}^i},s^{\boldsymbol{X}^i}) = Concat(head_1,...,head_n)W^O
\end{equation}

\begin{equation}
    \label{eq:cross_TFF_cross_multi}
    head_k = CrossAttention(f^{\boldsymbol{{I}}_{t_{1,2}}^i}W^Q_k,s^{\boldsymbol{X}^i}W^K_k,s^{\boldsymbol{X}^i}W^V_k)
\end{equation}
where $W^Q_k,W^K_k,W^V_k \in \mathbb{R}^{d_{E_I}\times d}$ are the learnable matrices of the $k^{th}$ head and $f^{\boldsymbol{{I}}_{t_{1,2}}^i}$ represents the bitemporal features in a compact manner. Following \cite{LiuCCD}, the output of the MultiHead attention mechanism is added to the original bitemporal patch embedding $f^{\boldsymbol{{I}}_{t_{1,2}}^i} \in \mathbb{R}^{T\times d_{E_I}}$
\begin{equation}
    \label{eq:cross_TFF_LN_1}
    f^{\boldsymbol{{'I}}_{t_{1,2}}^i} = LN(f^{\boldsymbol{{I}}_{t_{1,2}}^i}+  MultiHead(f^{\boldsymbol{{I}}_{t_{1,2}}^i}))
\end{equation}
\begin{equation}
    \label{eq:cross_TFF_cross_LN2}
    f^{\boldsymbol{{''I}}_{t_{1,2}}^i} = LN(f^{\boldsymbol{{'I}}_{t_{1,2}}^i}+  g(f^{\boldsymbol{{'I}}_{t_{1,2}}^i}))
\end{equation}
where $g(.)$ represents two fully connected layers with RELU activation function, respectively. The bitemporal features $f^{''I}_{t_{1,2}}$ (i.e., $f^{''I}_{t_1}$ and $f^{''I}_{t_2}$) obtained utilizing  (\ref{eq:cross_TFF_cross_LN2}) are then fused together through multiple fusion stages as depicted in Fig. \ref{Fig:bi_fusion}. A single fusion stage can be described by the following equation:
\begin{multline}
    \label{eq:cross_TFF_cross_LN3}
    f^{\boldsymbol{X}^i}_{l} = LN (Concat(f^{\boldsymbol{{''I}}_{t_{1}}^i},f^{\boldsymbol{{''I}}_{t_{2}}^i})+f^{\boldsymbol{X}^i}_{l-1} \\+ r(Concat(f^{\boldsymbol{{''I}}_{t_{1}}^i},f^{\boldsymbol{{''I}}_{t_{2}}^i})+ f^{\boldsymbol{X}^i}_{l-1}))   
\end{multline}
where $f^{\boldsymbol{X}^i}_{l-1} \in \mathbb{R}^{T\times 2d}$ represents the output of the $l-1$ fusion stage, $r(.)$ a residual block of three convolutional layers, batch normalization and drop out layers \cite{LiuCCD}.

\subsection{Network optimization}
Once the fused bitemporal images and text embeddings are obtained, they are passed through the respective modality-specific projection heads (see Fig. \ref{Fig:scheme}) that consist of feed-forward neural networks (FFN). These projection heads map the multimodal embeddings to a joint representation space of fixed dimensionality $d_F$
\begin{equation}
    \label{eq:img_project}
    F^{\boldsymbol{X}^i} = FNN_{X}(f^{\boldsymbol{X}^i})
\end{equation}
\begin{equation}    
    \label{eq:txt_project}
    F^{\boldsymbol{Y}^i} = FNN_{Y}(f^{\boldsymbol{Y}^i})
\end{equation}
where $F^{\boldsymbol{X}^i}\in {R}^{1\times d_{{F}}}$ and $F^{\boldsymbol{Y}^i}\in {R}^{1\times d_{{F}}}$ are the mapped bitemporal image and text representations, respectively. To learn the shared embedding space of text and bitemporal images, we employ the contrastive loss as described by Radford et al.\cite{Radford2021LearningTV}, which has proven effective for jointly aligning image and text representations using only paired data without requiring explicit labels. In detail, let $B = \{\boldsymbol{X}^i,\boldsymbol{Y}^i\}^b_{i=1}$ represent a mini-batch of size $b$ of randomly sampled bitemporal images and text pairs from the training set $\mathcal{D}$. Passing this mini-batch as input to the proposed model produces the following bitemporal images and text representations $\{F^{\boldsymbol{X}^i}\}^b_{i=1}$ and $\{F^{\boldsymbol{Y}^i}\}^b_{i=1}$, respectively. The learning objective is to maximize the similarity of $b$ correct bitemporal images and text pairs, while minimizing the similarity between $b^2-b$ incorrect pairs within the mini-batch \cite{Radford2021LearningTV}. To this end, the loss is computed in both the image and text domains. In the image domain, the objective is to bring the text features closer to their corresponding image features, while pushing them away from other text features within the mini-batch. Similarly, in the text domain, the aim is to bring the corresponding image features closer, while pushing away the other image features within the mini-batch. This is achieved through the bidirectional normalized temperature-scaled cross-entropy losses defined as follows \cite{sohn2016improved_NTX}:
\begin{equation}
    \label{eq:xy}
    L_{F^X \to F^Y} = -\frac{1}{b} \sum_{i=1}^{b} \log  \frac{\exp(cos(F^{\boldsymbol{X}^i},F^{\boldsymbol{Y}^i})\exp(\kappa))}{\sum_{j=1}^{b} \exp\left( cos(F^{\boldsymbol{X}^i},F^{\boldsymbol{Y}^j}) \exp(\kappa) \right)}
\end{equation}
\begin{equation}
    \label{eq:yx}
    L_{F^Y \to F^X} = -\frac{1}{b} \sum_{i=1}^{b} \log  \frac{\exp(cos(F^{\boldsymbol{Y}^i},F^{\boldsymbol{X}^i})\exp(\kappa))}{\sum_{j=1}^{b} \exp (cos(F^{\boldsymbol{Y}^i},F^{\boldsymbol{X}^j}) \exp(\kappa))}
\end{equation}
where $\kappa$ is a learnable temperature parameter that controls the sharpness of the distribution and $cos(\cdot)$ is the cosine similarity. To account for both contributions, the final loss is the average of the two loss terms:
\begin{equation}
    \label{eq:avg}
    \mathbf{}{L_C} = \frac{1}{2}(L_{F^X \to F^Y}+L_{F^Y \to F^X}).
\end{equation}
This final loss encourages symmetry and consistency in the learned space across modalities, enabling a robust joint representation for cross-modal ITSR.

After training, we obtain the features of bitemporal images and text sentences from the respective modality-specific projection heads. 
To retrieve semantically similar bitemporal images to a given a query text sentence $\boldsymbol{Y}^q$, we compute the cosine similarity between the query text sentence features and all bitemporal image features. 
The similarities are ranked in descending order, and the top-\textit{k} bitemporal images are retrieved. Similarly, when the query is provided as a bitemporal image $\boldsymbol{X}^q$, we compute its cosine similarity with 
all textual features, rank the results, and retrieve the top-\textit{k} text sentences.

\section{Dataset Description and Experimental Design} \label{dataset_setup}

\subsection{Datasets}

To assess the performance of the proposed method, we carried out our experiments on two benchmark archives. 

The first archive, LEVIR Change Captioning (LEVIR-CC) \cite{LiuCCD}, comprises 10077 RGB image pairs of size $256\times256$ pixels with a spatial resolution of 0.5 meters.  The image pairs within the LEVIR-CC dataset predominantly consider building-related changes in the area of Texas, USA. The acquisition time spans from 2002 to 2018, with a minimum of 5 and a maximum of 14 years between the acquisition of the two images of each pair. Each image pair is annotated with five different change descriptions. Notably, half of the dataset contains scenes depicting no change between the two images. In such cases, all the image pairs are annotated with the same set of 5 change captions.

The second archive, Dubai Change Captioning Dataset (Dubai CCD) \cite{HoxhaCCD} considers the urban development of the city of Dubai. It comprises 500  multispectral image pairs acquired by Enhanced Thematic Mapper Plus (ETM+) sensor onboard Landsat 7.  Each image pair captures a 10-year span, with the first image acquired on May 19, 2000 and the second on June 16, 2010. The bitemporal images are of size $50\times50$ pixels and contain 6 bands (i.e., R,G,B, near-infrared, short-wave infrared and mid-infrared) each of which is characterized by a spatial resolution of 30 meters. Each image pair is annotated with five different captions by expert annotators. Unlike the first archive, image pairs depicting no change are not always described by the same five captions. In such cases, various captions are provided, introducing variability even when there is no semantic difference between the images.

\subsection{Experimental Setup}
In our experiments, we used the predefined dataset splits to assess the performance of the proposed method. For the Dubai CCD  the splits are as follows: $60\%$, $10\%$, and $30\%$ for training, validation and testing, respectively. The splits for the LEVIR-CC dataset are $80\%$ for training, $10\%$ for validation, and $10\%$ for testing. Each caption is paired with its corresponding bitemporal image pair separately. To stabilize training for the LEVIR-CC dataset, we used only $15\%$ of the text and bitemporal image pairs that depict no changes. This decision was due to the dataset composition, where half of the samples represent no-change scenes annotated with the same set of five change descriptions.  For evaluation, we merged the test and validation splits employing a leave-one-out strategy where one example is used as the query and the rest is used as the archive. The evaluation is repeated five times, each time selecting a random caption for every image pair at each round. The final results are averaged across all rounds. The evaluation is done considering two retrieval tasks:  1) $T \rightarrow I$ is the task where the query is the text modality and the retrieval is applied to an archive of bitemporal images; and 2) $I \rightarrow T$ where the query is the image modality (i.e., bitemporal images) and the retrieval is applied to an archive  of text sentences. We utilized CLIP's pre-trained Vision Transformer (ViT-B/16) as image encoder \cite{Radford2021LearningTV}. In the TFF strategy we used a total of three fusion stages (i.e., $l=3$) and the patch embeddings as image representations. For the GFF strategies, we used the class token embedding as global features to represent the bitemporal images. We used CLIP's pre-trained text encoder and the class token embeddings to represent the text sentences \cite{Radford2021LearningTV}. We pass both scaled and channel-wise normalized images through the respective backbone networks before applying the fusion strategies.

The modal-specific feed-forward neural networks used in the projection heads consist of a hidden layer with 256 dimensions using ReLU activations and an output layer (i.e., $d_X, d_Y$) with 128 dimensions. The networks have a learnable temperature parameter $\kappa$ that is initialized to 0.07 as suggested in\cite{Radford2021LearningTV}. For optimization, we used mini-batch stochastic gradient descent (SGD) with a batch size of 32, a learning rate of 0.01, weight decay of $5 \times 10^{-4}$, and momentum of 0.9.
The models are trained for 30 epochs on NVIDIA A100 GPUs.

We compare our method against a baseline employing an early fusion strategy (EF), which is widely used in change detection problems. \cite{Baseline_method,FC_Conc_Differ_2018}. According to this strategy, the two images in each bitemporal pair are concatenated along the channel dimension to form a single input, without explicitly modeling their temporal relationship. To ensure a fair evaluation, we adopt the same experimental setup as in our proposed method, including the use of CLIP's Vision Transformer (ViT-B/16) as image encoder, the same pre-trained CLIP text encoder, identical projection heads, and training procedure.

\subsection{Evaluation Metrics}
We measured the retrieval performance using  bilingual evaluation understudy (BLEU) \cite{BLEU_metric}, metric for evaluation of translation with explicit ordering (METEOR) \cite{banerjee-lavie-2005-meteor} and recall-oriented understudy for Gisting evaluation (ROUGE-L)\cite{lin-2004-rouge}, for both $T \rightarrow I$ and $I \rightarrow T$ retrieval tasks. It is worth noting that the aforementioned metrics are commonly used in machine translation and in image captioning to measure the similarity between generated sentence descriptions (i.e., hypothesis) and the reference sentence descriptions. To this end, BLEU score measures the \textit{n}-gram (i.e., n-consecutive words) precision to quantify the similarity between the hypotheses and reference descriptions, where $n=1,4$ in this work. ROUGE-L is based on the calculation of the F-score with respect to the longest common subsequence between the hypothesis and the reference descriptions. Finally, METEOR computes the weighted F-score, with more weight given to the recall and considers semantic similarity of the words (i.e., synonyms). All the used metrics range from 0 to 1, where a score of 1 indicates a perfect match. 

To apply these metrics to our retrieval tasks, we appropriately defined the hypotheses and references. For the text-to-bitemporal image retrieval task (i.e.,  $T \rightarrow I$), the query text is treated as the hypothesis, while the reference captions of the retrieved bitemporal images served as references. In the bitemporal image-to-text retrieval task (i.e., $I \rightarrow T$), the five captions associated with the query bitemporal images are used as references, while the retrieved captions are considered as the hypothesis. This adaptation ensures consistent and fair evaluation in both retrieval tasks. We measure the retrieval performance using the aforementioned metrics on the top five retrieved data for each query. For a given query text or bitemporal images, the scores are computed individually for each retrieved data (i.e., bitemporal images or text) and then averaged. Finally, the overall performance is obtained averaging the scores for all the available queries.

\section{Experimental Results} \label{experimental_results}
In this section, we present the results of the proposed text-ITSR method with its different fusion strategies applied to two retrieval tasks: 1) $T \rightarrow I$;  and 2) $I \rightarrow T$ under three distinct scenarios: 1) full query set scenario (which includes queries, either text sentences or bitemporal images, that are associated to change and no change); 2) change query set scenario (which includes only those queries associated to change); and 3) no-change query set scenario (which includes only those textual or bitemporal images queries that are associated to no change). We also compare the proposed method against the EF baseline method.

\subsection{Experimental Results on Dubai CCD}

\begin{table*}[ht]
    \centering
    \caption{BLEU-1, BLEU-4, METEOR and ROUGE-L results on text-to-bitemporal images ($T\rightarrow I$), bitemporal image-to-text ($I\rightarrow T$) retrieval tasks and the average across the two retrieval tasks of the EF baseline and the proposed method under different fusion strategies on the DUBAI-CCD dataset using the \textbf{Full} query set.}
    \begin{tabular}{clccccccccccc}
    \hline
    \addlinespace
    Fusion Strategy &
    \multicolumn{3}{c}{BLEU-1} &
    \multicolumn{3}{c}{BLEU-4} & \multicolumn{3}{c}{METEOR} & \multicolumn{3}{c}{ROUGE-L} \\
    \cmidrule(lr){2-4} \cmidrule(lr){5-7} \cmidrule(lr){8-10} \cmidrule(lr){11-13}
        & I $\rightarrow$ T & T $\rightarrow$ I & Average
        & I $\rightarrow$ T & T $\rightarrow$ I & Average
        & I $\rightarrow$ T & T $\rightarrow$ I & Average
        & I $\rightarrow$ T & T $\rightarrow$ I & Average \\
    \hline
    \addlinespace
    EF Baseline & 0.55 & 0.533 & 0.541 & 0.262 & 0.205 & 0.229 & 0.273 & 0.249 & 0.261 & 0.468 & 0.462 & 0.465 \\
    GFF: Concatenation & 0.545 & 0.514 & 0.53 & 0.227 & 0.21 & 0.219 & 0.266 & 0.265 & 0.266 & 0.465 & 0.504 & 0.484 \\
    GFF: Subtraction & 0.569 & \textbf{0.607} & 0.588 & 0.236 & 0.28 & 0.258 & 0.277 & \textbf{0.322} & 0.304 & 0.479 & \textbf{0.551} & 0.515 \\
    TFF & \textbf{0.602} & \textbf{0.607} & \textbf{0.604} & \textbf{0.272} & \textbf{0.282} & \textbf{0.277} & \textbf{0.298} & 0.312 & \textbf{0.305} & \textbf{0.513} & 0.537 & \textbf{0.525} \\
    \hline
    \end{tabular}
    \label{tab:dubai-full}
\end{table*}

\begin{table*}[ht]
    \centering
    \caption{BLEU-1, BLEU-4, METEOR and ROUGE-L results on text-to-bitemporal images ($T\rightarrow I$), bitemporal image-to-text ($I\rightarrow T$) retrieval tasks and the average across the two retrieval tasks of the EF baseline and the proposed method under different fusion strategies on the DUBAI-CCD dataset using the \textbf{Change} query set.}
    \begin{tabular}{clccccccccccc}
    \hline
    \addlinespace
    Fusion Strategy &
    \multicolumn{3}{c}{BLEU-1} &
    \multicolumn{3}{c}{BLEU-4} & \multicolumn{3}{c}{METEOR} & \multicolumn{3}{c}{ROUGE-L} \\
    \cmidrule(lr){2-4} \cmidrule(lr){5-7} \cmidrule(lr){8-10} \cmidrule(lr){11-13}
        & I $\rightarrow$ T & T $\rightarrow$ I & Average
        & I $\rightarrow$ T & T $\rightarrow$ I & Average
        & I $\rightarrow$ T & T $\rightarrow$ I & Average
        & I $\rightarrow$ T & T $\rightarrow$ I & Average \\
    \hline
    \addlinespace
    EF Baseline & 0.462 & 0.537 & 0.499 & 0.149 & 0.169 & 0.159 & 0.195 & 0.238 & 0.217 & 0.369 & 0.430 & 0.400 \\
    GFF: Concatenation & \textbf{0.566} & 0.520 & 0.543 & \textbf{0.206} & 0.165 & 0.185 & \textbf{0.264} & 0.245 & 0.254 & \textbf{0.469} & 0.435 & 0.452 \\
    GFF: Subtraction & 0.527 & 0.560 & 0.544 & 0.179 & 0.201 & 0.190 & 0.238 & 0.274 & 0.256 & 0.431 & 0.471 & 0.451 \\
    TFF & 0.553 & \textbf{0.575} & \textbf{0.564} & 0.194 & \textbf{0.224} & \textbf{0.209} & 0.253 & \textbf{0.295} & \textbf{0.274} & 0.458 & \textbf{0.488} & \textbf{0.473} \\
    \hline
    \end{tabular}
    \label{tab:dubai-change}
\end{table*}

\begin{table*}[ht]
    \centering
    \caption{BLEU-1, BLEU-4, METEOR and ROUGE-L results on text-to-bitemporal images ($T\rightarrow I$), bitemporal image-to-text ($I\rightarrow T$) retrieval tasks and the average across the two retrieval tasks of the EF baseline and the proposed method under different fusion strategies on the DUBAI-CCD dataset using the \textbf{No-change} query set.}
    \begin{tabular}{clccccccccccc}
    \hline
    \addlinespace
    Fusion Strategy &
    \multicolumn{3}{c}{BLEU-1} &
    \multicolumn{3}{c}{BLEU-4} & \multicolumn{3}{c}{METEOR} & \multicolumn{3}{c}{ROUGE-L} \\
    \cmidrule(lr){2-4} \cmidrule(lr){5-7} \cmidrule(lr){8-10} \cmidrule(lr){11-13}
        & I $\rightarrow$ T & T $\rightarrow$ I & Average
        & I $\rightarrow$ T & T $\rightarrow$ I & Average
        & I $\rightarrow$ T & T $\rightarrow$ I & Average
        & I $\rightarrow$ T & T $\rightarrow$ I & Average \\
    \hline
    \addlinespace
    EF Baseline & \textbf{0.703} & 0.525 & 0.614 & \textbf{0.460} & 0.241 & 0.350 & \textbf{0.409} & 0.269 & 0.339 & \textbf{0.640} & 0.518 & 0.579 \\
    GFF: Concatenation & 0.508 & 0.505 & 0.507 & 0.264 & 0.289 & 0.277 & 0.271 & 0.300 & 0.285 & 0.457 & 0.622 & 0.540 \\
    GFF: Subtraction & 0.641 & \textbf{0.687} & 0.664 & 0.334 & \textbf{0.419} & 0.377 & 0.344 & \textbf{0.433} & \textbf{0.389} & 0.561 & \textbf{0.689} & \textbf{0.625} \\
    TFF & 0.686 & 0.662 & \textbf{0.674} & 0.408 & 0.382 & \textbf{0.395} & 0.376 & 0.341 & 0.358 & 0.609 & 0.624 & 0.617 \\
    \hline
    \end{tabular}
    \label{tab:dubai-nochange}
\end{table*}

\begin{figure}
    \centering
    \includegraphics[scale =0.35]{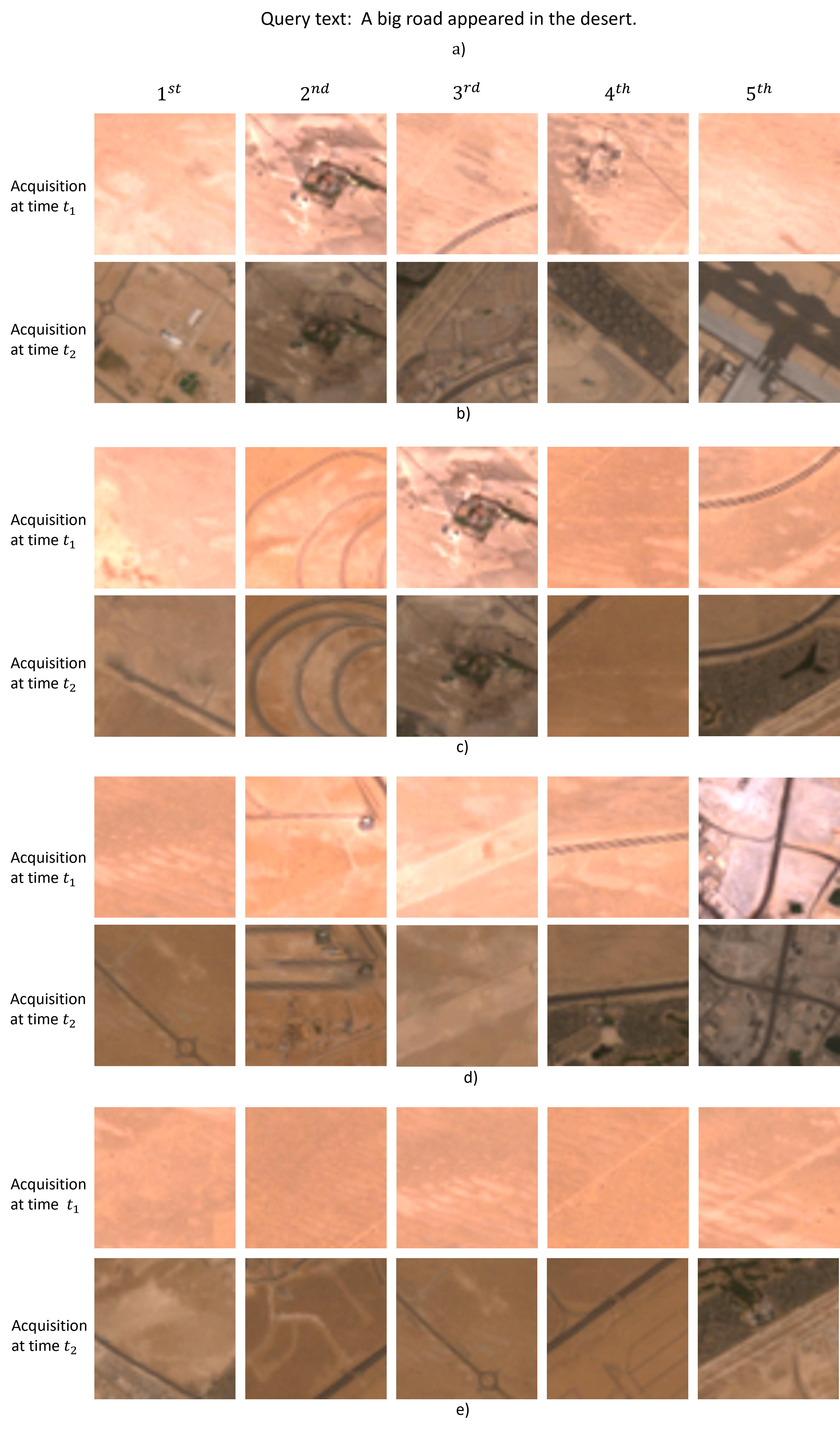}
    \caption{$T \rightarrow I$ retrieval example for Dubai CCD. a) query text. The retrieved bitemporal images by b) EF baseline method and our method when the: b) GFF: Concatenation, c) GFF: Subtraction and d) TFF strategies are used to model the semantic content of the bitemporal images.}

    \label{Fig:T-I DubaiCCD}
\end{figure}

\begin{figure*}
    \centering
    \includegraphics[scale =0.38]{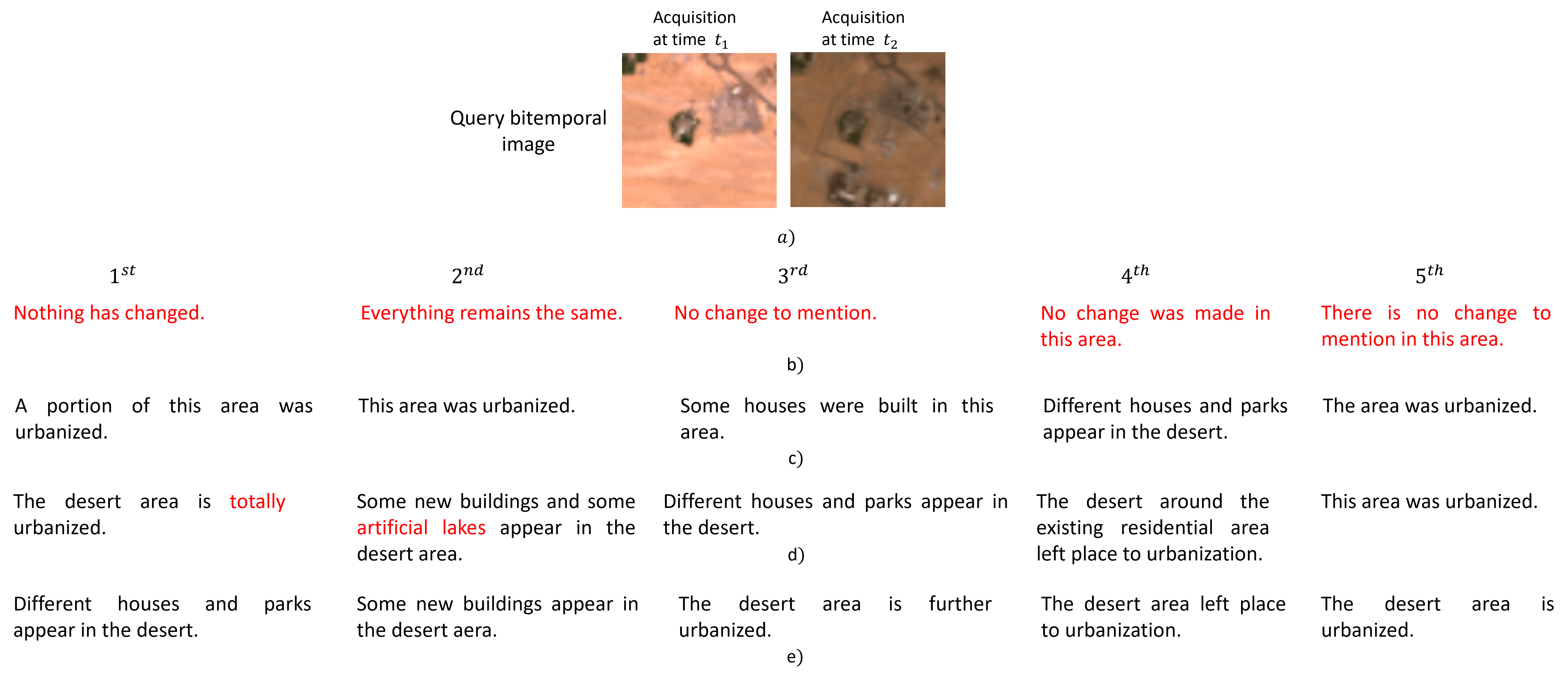}
    \caption{$I \rightarrow T$ retrieval example for Dubai CCD. a) query bitemporal images.  The retrieved text sentences by b) EF baseline method and our method when the: b) GFF: Concatenation, c) GFF: Subtraction and d) TFF strategies are used to model the semantic content of bitemporal images. Words highlighted in red indicate minor discrepancies or errors in the retrieved descriptions compared to the changes present in the
    query bitemporal image.}

    \label{Fig:I-T DubaiCCD}
\end{figure*}

\subsubsection{Full query set scenario} Table ~\ref{tab:dubai-full} \ shows the retrieval performances of the EF baseline method and our method under different fusion strategies on the full query set scenario. From the table one can see that the proposed method achieves the highest performance in the $T\rightarrow I$ retrieval task when using either the TFF or GFF: Subtraction strategy to model the semantic content of bitemporal images. For instance, in terms of BLEU-4 our method achieves a score of 0.28 and 0.282 using the TFF and GFF: Subtraction strategies. This corresponds to improvements of approximately $7 \%$ over the GFF: Concatenation strategy and $8 \%$ relative to the EF baseline. In other words, the reference sentence descriptions of bitemporal images retrieved by our method when utilizing the TFF or GFF: Subtraction strategy contain the highest number of overlapping 4-grams (i.e., 4 consecutive words) with the query sentence description. This indicates that the retrieved bitemporal images utilizing these two strategies in our method are highly correlated with the query sentence description. This performance behavior is further confirmed by looking at the  scores of METEOR and ROUGE-L. 

For the $I\rightarrow T$ retrieval task, one can notice that with the TFF strategy the proposed method achieves the highest retrieval scores across all the metrics. As an example, with the TFF strategy the proposed method achieves a ROUGE-L score of 0.513, outperforming the utilization of the GFF strategies and the EF baseline by approximately $4 \%$. When analzying the average retrieval scores across the two tasks, one can notice that with the TFF strategy once again the proposed method achieves the highest retrieval results. These results highlights the capability of the TFF strategy in modeling the semantic content of bitemproal images. As a result with the TFF strategy our method achieves the best retrieval performance across both tasks. Among the GFF variants, Subtraction outperforms Concatenation, due to its ability to explicitly model the semantic differences between bitemporal images. In contrast, the EF baseline and our method with GFF: Concatenation exhibit similarly poor performance, as they lack of mechanisms to highlight semantic changes. We further analyze change-specific and no-change query sets to assess whether the observed performances are consistent. 

\subsubsection{Change query set scenario} Table ~\ref{tab:dubai-change} shows the retrieval performances of the EF baseline method and our method under different fusion strategies on the change set scenario. Also in this scenario, one can notice that when the TFF strategy is used to model the semantic content of the bitemporal images, our method achieves the highest average retrieval scores across both retrieval tasks and all evaluation metrics.For instance, in the $T \rightarrow I$ retrieval task, our method with the TFF strategy achieves a BLEU-4 score of 0.224, outperforming both the GFF–Concatenation variant and the EF baseline by approximately $6 \%$. With the TFF our method shows an improvement of  $2 \%$ over the GFF: Subtraction.  A similar trend is observed for METEOR and ROUGE-L, further confirming the capability of TFF to effectively capture semantic changes in bitemporal images. This enables the proposed method to achieve accurate alignment between bitemporal images and textual descriptions that are associated with change, resulting in the highest retrieval accuracy in the $T\rightarrow I$ task.
For the $I \rightarrow T$ retrieval task, the highest retrieval scores by our method are obtained using the GFF: Concatenation strategy. For instance, with the GFF: Concatenation  our method achieves a BLEU-4 score of 0.206, a METEOR score of 0.264, and a ROUGE-L score of 0.469, showing a slight improvement (up to $1 \%$ ) over the utilization of the TFF strategy. Compared to the GFF: Subtraction strategy, these improvements increase to approximately $3\%$. In this retrieval task, the lowest results are again achieved by the EF baseline method, showing its limited ability to represent semantic changes and align them with the textual descriptions. 

When considering the average performances across both tasks, the integration of the TFF strategy into our method yields the highest retrieval results, making it the best choice to model the semantic content of bitemporal images associated with change. Specifically, the integration of the TFF strategy provides an improvement over the GFF strategies of around $2 \%$ in all the considered metrics. More notably with the TFF strategy our method outperforms the EF baseline by a wider margin ( approximately $5 \%$ in BLEU-4) highlighting the limitations of EF that fails to explicitly model changes over the bitemporal images. These results further emphasize the robustness of the TFF strategy in modeling changes over bitemporal images. Thus, when integrated into the proposed text-ITSR method, it allows more accurate retrieval compared to both the GFF strategies and the EF baseline method.

\subsubsection{No-change query set scenario} Table ~\ref{tab:dubai-nochange}  shows the retrieval performances of the EF baseline method and our method under different fusion strategies on the change set scenario. In this scenario, we observe a distinct behavior compared to previous scenarios. While for the $T\rightarrow I$ retrieval task our method achieves the highest retrieval performance, for the $I\rightarrow T$ retrieval task, the EF baseline method  yields the best results. For instance, for $T\rightarrow I$ retrieval task, with the TFF strategy our method achieves a BLEU-4 score of 0.382 and a METEOR score of 0.341 representing an improvement of approximately $14 \%$ and $8 \%$ over the EF baseline method. In contrast, for $I\rightarrow T$ retrieval task, the EF baseline method achieves a BLEU-4 score of 0.460 and a METEOR score of 0.409 outperforming our method with TFF by approximately $6 \%$ and $3 \%$.  This can be explained by considering the nature of the dataset and the evaluation metrics. In Dubai CCD, even bitemporal images associated with no-change, are annotated with multiple textual descriptions, with some occurring more frequently than others. The TFF strategy is explicitly designed to learn change representations from bitemporal image pairs. Consequently, with TFF our method often retrieves valid textual descriptions associated with no-change that are less frequent or linguistically different from the reference descriptions of the query. As a result, it can be penalized by n-gram-based metrics like BLEU and ROUGE-L, which favor exact word matches. In contrast, EF tends to retrieve more frequent textual descriptions, which increases overlap with reference captions and thus improves n-gram-based metrics like BLEU and ROUGE-L. This discrepancy is less pronounced with METEOR, which is more tolerant to linguistic variation and synonymy. The relatively small METEOR score gap ($3 \%$) between EF and our TFF-based method supports this interpretation.  Furthermore, the EF baseline exhibits greater performance asymmetry across retrieval directions. As shown in Table \ref{tab:dubai-nochange}, the METEOR score difference between $T\rightarrow I$ and $I\rightarrow T$ for the EF baseline is of $13 \%$ whereas for TFF-based method is only $2 \%$. This indicates that the integration of TFF in our method yields more consistent, balanced and overall higher performance across retrieval tasks further showing its robustness for text-ITSR. 

Regarding the use of the GFF variants, we observe that the integration of the GFF: Subtraction in our method is consistently more effective than that of GFF: Concatenation across the two retrieval tasks. Specifically, with the GFF: Subtraction our method performs best on $T\rightarrow I$ retrieval task, achieving a BLEU-4 score of 0.419, outperforming the integration of the TFF and GFF: Concatenation by $3 \%$ and $12 \%$, respectively. It also performs better than the GFF: Concatenation in the $I\rightarrow T$ task. However, in this task with the TFF our method yields the highest retrieval performance (e.g, BLEU-4 = 0.408), surpassing both GFF: Concatenation and GFF: Subtraction by  $13 \%$ and $7 \%$.  By analyzing Table \ref{tab:dubai-nochange}, one can notice that using the TFF strategy in the proposed method results in obtaining the highest scores across the two tasks. As an example, with the TFF strategy the proposed method attains an average BLEU-4 score of 0.395 which is approximately $10\%$ and $6 \%$ higher compared to the one achieved when GFF: Concatenation and GFF: Subtraction is integrated, respectively. This shows that the TFF strategy is more robust compared to the other two strategies in modeling the semantic content over bitemporal images. 

Figure \ref{Fig:T-I DubaiCCD} shows a $T\rightarrow I$ retrieval example where the query text is: \textit{"A big road appeared in the desert."} and the bitemporal images retrieved by the EF baseline method [see Fig. \ref{Fig:T-I DubaiCCD} b)] and the proposed  method using the GFF: Concatenation [Fig. \ref{Fig:T-I DubaiCCD} c)], GFF: Subtraction [see Fig. \ref{Fig:T-I DubaiCCD} d)] and TFF [see Fig. \ref{Fig:T-I DubaiCCD} e)] strategies to model the semantic content of the bitemporal images. One can notice that most of the retrieved bitemporal images by our method correctly depict changes involving the appearance of a road in the desert. In contrast, the retrieved bitemporal images by the EF baseline method are less accurate. A closer inspection reveals that the retrieved images by the proposed method with the TFF strategy exhibit the highest semantic alignment with the query text. Specifically, the first four retrieved bitemporal images by our method when utilizing the TFF strategy depict exactly the appearance of a road in a desert. On the other hand, when utilizing the GFF strategies some inaccuracies are observed in the retrieved images.  For instance, the $3^{rd}$ and $5^{th}$ retrieved images do not align with the query text sentence. This is even more obvious on the bitemporal images retrieved by the EF baseline method where only one of them depicts the appearance of a road in the desert ($2^{nd}$ retrieved bitemporal image). The rest of them depict broader changes related to the urbanization of the area.

Figure \ref{Fig:I-T DubaiCCD} shows an $I\rightarrow T$ retrieval example, where one of the reference descriptions of the query bitemporal images [Fig. \ref{Fig:I-T DubaiCCD} a) ] is \textit{"Many new buildings and new roads appear in the desert."} and the text sentences retrieved by the EF baseline method [Fig. \ref{Fig:I-T DubaiCCD} b)] and the proposed method using the GFF: Concatenation [Fig. \ref{Fig:I-T DubaiCCD} c)], GFF: Subtraction [see Fig. \ref{Fig:I-T DubaiCCD} d)] and TFF [see Fig. \ref{Fig:I-T DubaiCCD} e)] strategies. Notably, all descriptions retrieved by the proposed method successfully capture the change between the two image acquisitions that is "the presence of new buildings in the desert area". In contrast, the EF baseline method is unable to retrieve any relevant descriptions to the query bitemporal image. Specifically, all the descriptions retrieved by the EF baseline method are associated with no change, highlighting its limitation in modeling the semantic changes. One can notice that the retrieved descriptions by the proposed method are all semantically relevant to the query bitemporal images. In particular, the retrieved descriptions by the proposed method with the TFF show a stronger semantic alignment with the query bitemporal image. They effectively describe that the area has undergone urban development or that new buildings have appeared, demonstrating the strength of TFF in effectively capturing semantic changes between bitemporal images.

\subsection{Experimental Results on LEVIR CC}

\begin{table*}[ht]
    \centering
    \caption{BLEU-1, BLEU-4, METEOR, and ROUGE-L results on text-to-bitemporal images ($T\rightarrow I$), bitemporal image-to-text ($I\rightarrow T$), and the average across the two retrieval tasks of the EF baseline and our method under different fusion strategies on the LEVIR-CC dataset using the \textbf{Full} query set.}
    \begin{tabular}{clccccccccccc}
    \hline
    Fusion Strategy &
    \multicolumn{3}{c}{BLEU-1} &
    \multicolumn{3}{c}{BLEU-4} & \multicolumn{3}{c}{METEOR} & \multicolumn{3}{c}{ROUGE-L} \\
    \cmidrule(lr){2-4} \cmidrule(lr){5-7} \cmidrule(lr){8-10} \cmidrule(lr){11-13}
        & I $\rightarrow$ T & T $\rightarrow$ I & Average
        & I $\rightarrow$ T & T $\rightarrow$ I & Average
        & I $\rightarrow$ T & T $\rightarrow$ I & Average
        & I $\rightarrow$ T & T $\rightarrow$ I & Average \\
    \hline
    EF Baseline & 0.524 & 0.764 & 0.644 & 0.249 & 0.565 & 0.407 & 0.316 & \textbf{0.606} & 0.461 & 0.441 & 0.688 & 0.565 \\
    GFF: Concatenation  & 0.597 & 0.698 & 0.647 & 0.331 & 0.487 & 0.409 & 0.388 & 0.407 & 0.397 & 0.512 & 0.616 & 0.564 \\
    GFF: Subtraction & 0.577 & 0.648 & 0.612 & 0.313 & 0.434 & 0.373 & 0.362 & 0.355 & 0.358 & 0.492 & 0.575 & 0.534 \\
    TFF & \textbf{0.712} & \textbf{0.775} & \textbf{0.744} & \textbf{0.477} & \textbf{0.568} & \textbf{0.523} & \textbf{0.515} & 0.565 & \textbf{0.540} & \textbf{0.625} & \textbf{0.695} & \textbf{0.660} \\
    \hline
    \end{tabular}
    \label{tab:LEVERCC_full}
\end{table*}

\begin{table*}[ht]
    \centering
    \caption{BLEU-1, BLEU-4, METEOR, and ROUGE-L results on text-to-bitemporal images ($T\rightarrow I$), bitemporal image-to-text ($I\rightarrow T$), and the average across the two retrieval tasks of the EF baseline and our method under different fusion strategies on the LEVIR-CC dataset using the \textbf{Change} query set.}
    \begin{tabular}{clccccccccccc}
    \hline
    Fusion Strategy &
    \multicolumn{3}{c}{BLEU-1} &
    \multicolumn{3}{c}{BLEU-4} & \multicolumn{3}{c}{METEOR} & \multicolumn{3}{c}{ROUGE-L} \\
    \cmidrule(lr){2-4} \cmidrule(lr){5-7} \cmidrule(lr){8-10} \cmidrule(lr){11-13}
        & I $\rightarrow$ T & T $\rightarrow$ I & Average
        & I $\rightarrow$ T & T $\rightarrow$ I & Average
        & I $\rightarrow$ T & T $\rightarrow$ I & Average
        & I $\rightarrow$ T & T $\rightarrow$ I & Average \\
    \hline
    EF Baseline & 0.573 & 0.529 & 0.551 & 0.135 & 0.131 & 0.133 & 0.221 & 0.212 & 0.216 & 0.402 & 0.377 & 0.389 \\
    GFF: Concatenation & 0.594 & 0.576 & 0.585 & 0.166 & 0.169 & 0.167 & 0.235 & 0.232 & 0.233 & 0.423 & 0.416 & 0.420 \\
    GFF: Subtraction & 0.591 & 0.538 & 0.564 & 0.166 & 0.147 & 0.156 & 0.232 & 0.219 & 0.225 & 0.420 & 0.393 & 0.406 \\
    TFF & \textbf{0.613} & \textbf{0.600} & \textbf{0.607} & \textbf{0.189} & \textbf{0.186} & \textbf{0.187} & \textbf{0.242} & \textbf{0.241} & \textbf{0.242} & \textbf{0.438} & \textbf{0.432} & \textbf{0.435} \\
    \hline
    \end{tabular}
    \label{tab:LEVERCC_change}
\end{table*}

\begin{table*}[ht]
    \centering
    \caption{BLEU-1, BLEU-4, METEOR, and ROUGE-L results on text-to-bitemporal images ($T\rightarrow I$), bitemporal image-to-text ($I\rightarrow T$), and the average across the two retrieval tasks of the EF baseline and our method under different fusion strategies on the LEVIR-CC dataset using the \textbf{No-change} query set.}
    \begin{tabular}{clccccccccccc}
    \hline
    Fusion Strategy &
    \multicolumn{3}{c}{BLEU-1} &
    \multicolumn{3}{c}{BLEU-4} & \multicolumn{3}{c}{METEOR} & \multicolumn{3}{c}{ROUGE-L} \\
    \cmidrule(lr){2-4} \cmidrule(lr){5-7} \cmidrule(lr){8-10} \cmidrule(lr){11-13}
        & I $\rightarrow$ T & T $\rightarrow$ I & Average
        & I $\rightarrow$ T & T $\rightarrow$ I & Average
        & I $\rightarrow$ T & T $\rightarrow$ I & Average
        & I $\rightarrow$ T & T $\rightarrow$ I & Average \\
    \hline
    EF Baseline & 0.475 & \textbf{1.000} & 0.738 & 0.363 & \textbf{1.000} & 0.681 & 0.410 & \textbf{1.000} & 0.705 & 0.480 & \textbf{1.000} & 0.740 \\
    GFF: Concatenation & 0.600 & 0.820 & 0.710 & 0.496 & 0.805 & 0.651 & 0.541 & 0.582 & 0.562 & 0.602 & 0.815 & 0.708 \\
    GFF: Subtraction & 0.563 & 0.757 & 0.660 & 0.459 & 0.721 & 0.590 & 0.491 & 0.491 & 0.491 & 0.565 & 0.757 & 0.661 \\
    TFF & \textbf{0.811} & 0.951 & \textbf{0.881} & \textbf{0.765} & 0.951 & \textbf{0.858} & \textbf{0.787} & 0.889 & \textbf{0.838} & \textbf{0.812} & 0.959 & \textbf{0.885} \\
    \hline
    \end{tabular}
    \label{tab:LEVERCC_nochange}
\end{table*}

\subsubsection{Full query set scenario} Table ~\ref{tab:LEVERCC_full}  shows the retrieval performances of the EF baseline method and our method under different fusion strategies on the full query set scenario. From the table one can see that the proposed method achieves the highest retrieval performance for both $T \rightarrow I$ and $I \rightarrow T$ retrieval tasks when employing the TFF strategy to model the semantic content of bitemporal images.
Specifically, for the $T \rightarrow I$ retrieval task, the proposed method achieves a BLEU-4 score of 0.568 using the TFF strategy, outperforming the use of  GFF: Subtraction and GFF: Concatenation strategies by 13$\%$ and 8$\%$, respectively. This superior performance is further confirmed by looking at the achieved scores in terms of METEOR and ROUGE-L. For instance, our method achieves a METEOR score of 0.565 with the TFF strategy, which is 21$\%$ and 16$\%$ higher than the scores achieved with the GFF: Subtraction and GFF: Concatenation strategies, respectively. Similarly, a ROUGE-L score of 0.695 is obtained when using the TFF strategy, representing an improvement of 12$\%$ and 8$\%$ over the GGF: Subtraction and GFF concatenation strategies, respectively.  Consequently, integrating the TFF strategy into our method significantly enhances its ability to retrieve semantically aligned bitemporal images for query sentence descriptions, outperforming the GFF strategies.

For the $I \rightarrow T$ retrieval task, the TFF-based method continues to achieve the highest retrieval performance. Specifically, it obtains a BLEU-4 score of 0.477, surpassing GFF: Subtraction and GFF: Concatenation by 16$\%$ and 14$\%$, respectively. This indicates that the retrieved textual descriptions by our method using the TFF are more semantically aligned with the query bitemporal images compared to those retrieved with the GFF strategies. The EF baseline method performs comparably to our method with the TFF in $T \rightarrow I$ retrieval but shows significantly lower performance in $I \rightarrow T$, with a BLEU-4 score of 0.249. This score is 22$\%$ and approximately 10$\%$ lower than the proposed method with TFF strategy and the GFF variants, respectively. These results show that EF baseline method is unable to effectively model the semantic content of bitemporal images compared to our fusion strategies. As a result, the EF baseline is less effective for the $I \rightarrow T$ retrieval task compared to $T \rightarrow I$ retrieval task. This is because bitemporal images exhibit greater semantic variability compared to textual sentences within this dataset, making $I \rightarrow T$ more challenging compared to $T \rightarrow I$ retrieval task. In the following, we analyze the change and no-change query sets separately to better understand these behaviors.

 \begin{figure}
    \centering
    \includegraphics[scale =0.35]{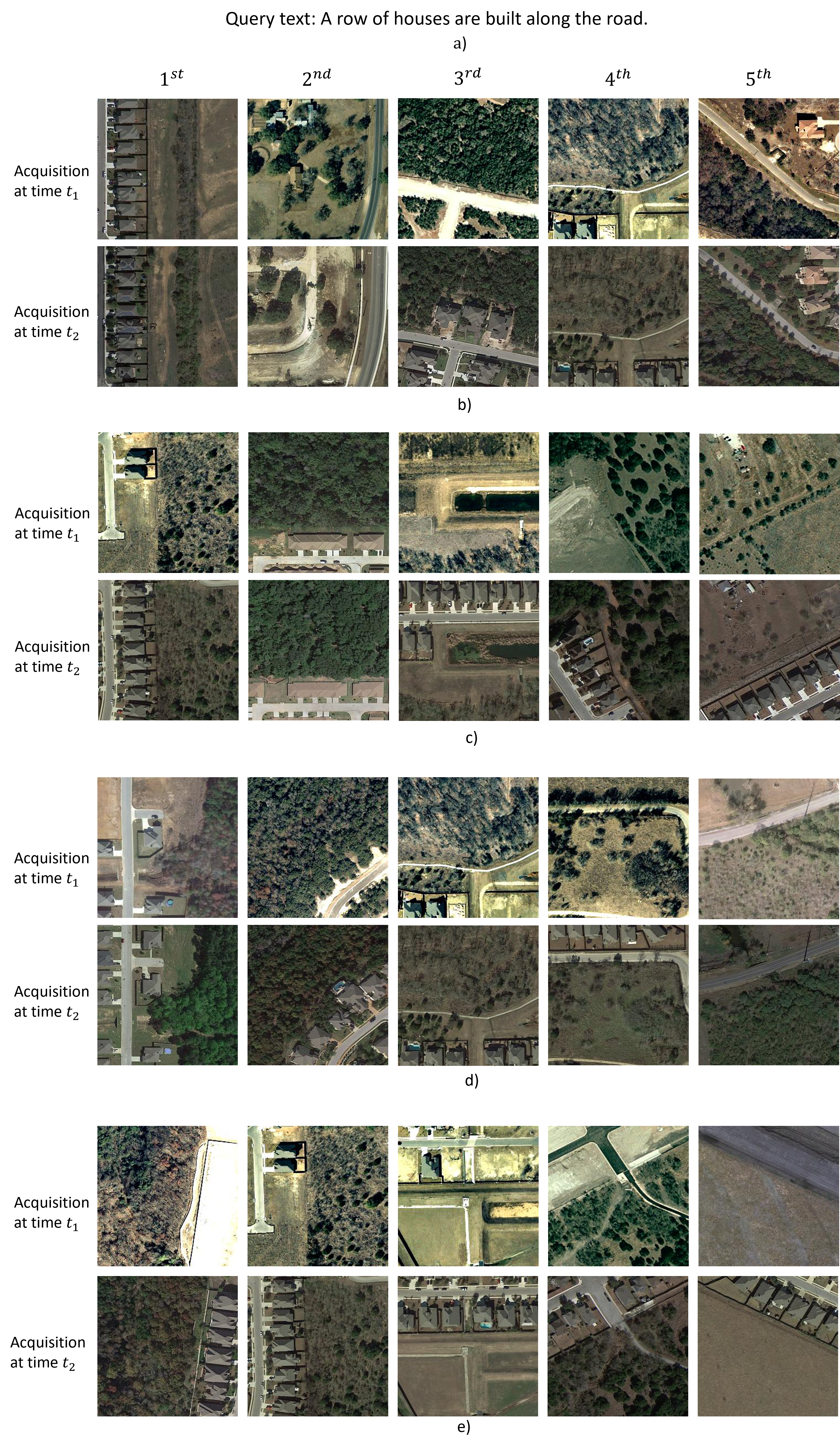}
    \caption{$T \rightarrow I$ retrieval example for LEVIR CC. a) query text and the retrieved bitemporal images by b) the EF baseline method and our method when the: c) GFF: Concatenation, d) GFF: Subtraction and e) TFF strategies are used to model the semantic content of bitemporal images.}

    \label{Fig:T-I LEVIRCC_big}
\end{figure}

\begin{figure*}
    \centering
    \includegraphics[scale =0.38]{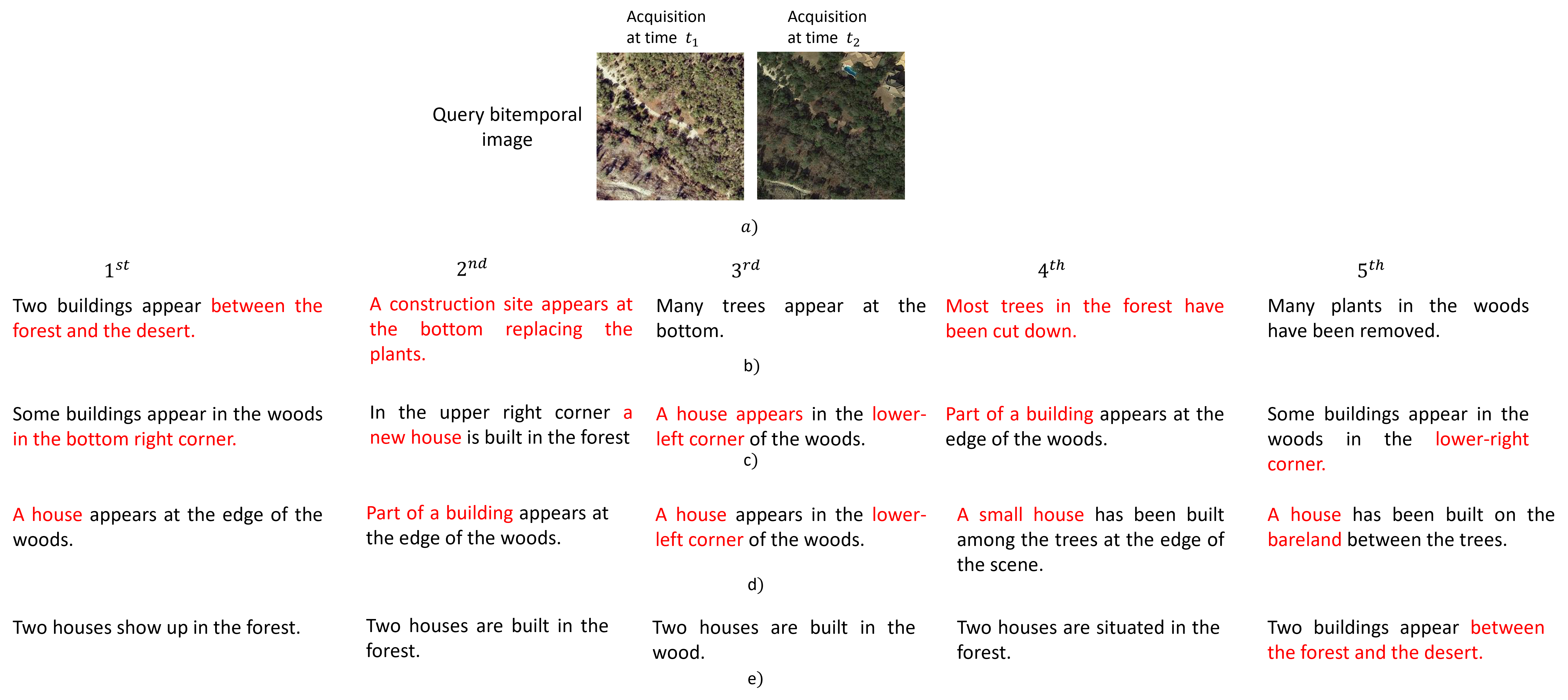}
    \caption{$I \rightarrow T$ retrieval example for LEVIR CC. a) query bitemporal images and retrieved text sentences by b) EF baseline method and our method when: c) GFF: Concatenation, d) GFF: Subtraction and e) TFF strategies are used to model the semantic content of bitemporal images. Words highlighted in red indicate minor discrepancies or errors in the retrieved descriptions compared to the changes present in the query bitemporal image.}

    \label{Fig:I-T LEVIRCC_big}
\end{figure*}

\subsubsection{Change query set scenario} Table ~\ref{tab:LEVERCC_change} shows the retrieval performances of the EF baseline method and our method under different fusion strategies on the change query set scenario. Consistent with the results from the full query set, the proposed method achieves the highest retrieval scores across all metrics and both retrieval tasks when employing the TFF strategy to model the semantic content of bitemporal images. For the $T \rightarrow I$ retrieval task, the proposed method with the TFF strategy achieves a BLEU-4 score of 0.186, surpassing the utilization of the GFF: Subtraction and GFF: Concatenation strategies by approximately $4 \%$ and $2 \%$, respectively. Moreover, our method consistently outperforms also the EF baseline regardless of the chosen fusion approach. Specifically, it achieves improvements of $1.5 \%$ with the GFF: Concatenation, $4 \%$ with GFF: Subtraction, and up to $5.5\%$ with TFF strategy. These results highlight the limitations of EF baseline on the change query set scenario where simply stacking the bitemporal images at the input level fails to capture temporal changes effectively. A similar trend is observed for the METEOR and ROUGE-L metrics. This improvement can be attributed to the TFF strategy's ability to learn discriminative features that emphasize temporal changes, which are crucial for retrieving bitemporal images that accurately reflect the textual descriptions of change.

In the $I \rightarrow T$ retrieval task, we can similarly observe that with the TFF strategy the proposed method continues to achieve higher results compared to the utilization of the GFF strategies and the EF baseline method. For instance, with the TFF strategy the proposed method achieves a BLEU-4 score of 0.189, outperforming the utilization of the GFF: variants by up to $3.3 \%$ and the EF baseline by $5.5 \%$. Similarly, METEOR and ROUGE-L metrics further demonstrate TFF's strong performance in the $I \rightarrow T$ retrieval task. These results emphasize the robustness of the TFF in handling changed-focused queries across modalities, thanks to its capabilities to explicitly model bitemporal images semantic content. This is further observed by analyzing the average scores across both  retrieval tasks where the TFF strategy clearly outperforms the GFF strategies and the EF baseline. Overall, these findings emphasize the need for dedicated fusion mechanisms that can effectively model semantic changes across bitemporal images for accurate cross-modal text-ITSR.

\subsubsection{No-change query set scenario} Table ~\ref{tab:LEVERCC_nochange} shows the retrieval performances of the EF baseline method and our method under different fusion strategies on the no-change query set scenario. From the table one can notice that for $T \rightarrow I$ retrieval task, the utilization of the TFF strategy continues to dominate, by achieving the highest scores across all metrics, significantly outperforming the utilization of  GFF strategies. Specifically, with the TFF strategy the proposed method achieves an almost perfect score (0.951), significantly outperforming the utilization of the GFF strategies by approximately $23 \%$, and $ 15 \%, $ in terms of BLEU-4 score for Subtraction and Concatenation variants, respectively. Similar behavior is also observed in the METEOR and ROUGE-L scores. Notably, the EF baseline also shows a very high performance in the $T \rightarrow I$ task, achieving perfect scores (1.000) across all metrics. However, this result is not indicative of a robust capability to model temporal semantics. Rather, it is a consequence of a structural peculiarity in the LEVIR-CC dataset: the no-change query set contains only five distinct sentences that describe the no-change scenarios. The retrieval set contains 2000 bitemporal images, half of which associated with no-change. These five sentences are used as queries to retrieve relevant images from a pool of 2000 bitemporal image pairs, of which half correspond to no-change scenarios.  Most likely, the EF baseline learns to distinguish low-variance (i.e., no-change) patterns, enabling it to retrieve visually similar images (i.e., no-change bitemporal images) without explicitly modeling temporal semantic. By contrast, the $I \rightarrow T$ retrieval is a substantially more fine-grained and challenging task. In this task, each of the 1000 no-change bitemporal image pairs is used as a separate query. Although these image pairs depict no change, they are diverse in terms of semantic content (i.e., urban area, forests, streets etc). This diversity means that a simple shortcut, such as detecting low variance between bitemporal images, is insufficient to accurately characterize no-change scenarios. Meanwhile, the retrieval set contains 5005 textual descriptions, but only five correspond to no-change cases, making this retrieval task (i.e., $I \rightarrow T$) highly imbalanced. In this  $I \rightarrow T$ task, successful retrieval requires not only modeling the temporal dynamics (i.e., distinguishing changed areas from no-change) but also capturing the semantic content of the bitemporal images to enable accurate alignment with the textual descriptions. Since the EF lacks an explicit modeling of both temporal and diverse semantic content in bitemporal images, it struggles to learn an effective alignment of the visually diverse no-change bitemporal images and textual descriptions. As a result, EF achieves the lowest BLEU-4 score (0.363) in this task. By contrast, the proposed method with TFF explicitly captures both the temporal consistency and the diverse semantic content of the no-change bitemporal images, producing representative features that enable more accurate alignment with the corresponding textual descriptions.  With the TFF strategy the proposed method achieves a BLEU-4 score of 0.765, significantly outperforming the utilization of the GFF: Subtraction by $30 \%$,  Concatenation strategies by approximately $26 \%$ and EF baseline by $40 \%$. A similar trend is observed in the METEOR and ROUGE-L scores, demonstrating the robustness of the proposed method with TFF strategy in retrieving no-change captions even under such a strong retrieval imbalance.

Fig. \ref{Fig:T-I LEVIRCC_big} shows a $T\rightarrow I$ retrieval example where the query text is: \textit{"A row of houses are built along the road."} and the bitemporal images retrieved by the EF baseline method and the proposed cross-modal text-ITSR method using the GFF: Concatenation [Fig. \ref{Fig:T-I LEVIRCC_big} c)], GFF: Subtraction [see Fig. \ref{Fig:T-I LEVIRCC_big} d)] and TFF [see Fig. \ref{Fig:T-I LEVIRCC_big} e)] strategies to model the semantic content of the bitemporal images. First, one can notice that most of the retrieved bitemporal images by our method depict changes involving the construction of rows of houses on the sides of the road. In contrast, those retrieved by the EF baseline method are less semantically aligned with the query text. Upon a closer examination, the retrieved bitemporal images when using the TFF strategy in the proposed method exhibit the strongest alignment with the query text. Specifically, with the TFF strategy [Fig. \ref{Fig:T-I LEVIRCC_big}(e)], all the retrieved bitemporal images represent changes involving the construction of a row of houses along the road. While with the GFF strategies the proposed method retrieves images depicting general construction activities along a road, they mostly fail to retrieve images that strictly match the "a single row of houses" criterion. For instance, with the GFF: Subtraction, only the $2^{nd}$ and $4^{rth}$ bitemporal images matches the query text precisely, while the $5^{th}$ retrieved bitemporal image shows no change at all. With the GFF: Concatenation strategy, the proposed method performs slightly better than with GFF: Subtraction. However, the $2^{nd}$ and $3^{rd}$ retrieved images deviate from the query text, depicting, respectively, the addition of a building within an existing residential area and the construction of two rows of houses along a road.

As shown in Fig. \ref{Fig:T-I LEVIRCC_big} b), the EF baseline method contains the highest number of retrieved bitemporal images which is less aligned with the query text. None of the retrieved bitemporal images fully matches the semantic content of the query.  This limitation is particularly evident in the first retrieved bitemporal image, which demonstrates a clear failure to model semantic change: both acquisitions depict an identical scene in which a row of houses is already present along the road, indicating the absence of any construction-related change. Nevertheless, the EF baseline incorrectly considers this pair relevant to the query text, which explicitly refers to the construction of a row of houses. Furthermore, $2^{nd}$ retrieved bitemporal image depicts changes related to the removal of a house, showing that the EF baseline is unable to distinguish between 'before' and 'after' in the bitemporal images. While the rest of the retrieved bitemporal images by the EF baseline method show the appearance of buildings along the road they do not match the "a single row of houses" criterion. 

These qualitative results highlight the superiority of the TFF strategy in effectively modeling the changes within bitemporal images compared to the GFF strategies and the EF baseline method. Consequently, the use of the TFF strategy in the proposed text-ITSR method achieves better alignment between the textual and image modalities, leading to more accurate $T \rightarrow I$ retrieval results.

Fig. \ref{Fig:I-T LEVIRCC_big} presents an $I\rightarrow T$ retrieval example where two of the reference descriptions of the query bitemporal images [Fig. \ref{Fig:I-T LEVIRCC_big} a) ] is \textit{"Two houses have been built at the corner of the scene."} and \textit{"Two houses are built in the forest."} The figure presents the top retrieved textual descriptions retrieved by the EF baseline [Fig. \ref{Fig:I-T LEVIRCC_big} b)] and the proposed cross-modal text-ITSR method using the GFF: Concatenation [Fig. \ref{Fig:I-T LEVIRCC_big} c)], GFF: Subtraction [see Fig. \ref{Fig:I-T LEVIRCC_big} d)] and TFF [see Fig. \ref{Fig:I-T LEVIRCC_big} e)] strategies to model the semantic content of the bitemporal images. One can see that almost all the retrieved textual descriptions by our method semantically correspond to the change between the two image acquisitions: the appearance of building/houses in the forest. In contrast, the retrieved textual descriptions by the EF baseline method are less semantically aligned with the bitemporal image.
One can notice that using TTF, the retrieved text descriptions by our method exhibit the strongest semantic alignment with the query bitemporal images. In particular, with the TFF strategy all the retrieved textual descriptions [see Fig. \ref{Fig:I-T LEVIRCC_big} e)] correctly describe changes involving the appearance of two houses in the forest. While both GFF strategies enable the retrieval of descriptions referring to the appearance of houses, they often fail to capture the correct number or spatial location.  For example, the $1^{st}$ retrieved sentence with the GFF: Concatenation strategy describes the appearance of the buildings in the woods at the bottom right corner.  However, in the query bitemporal image the two houses appear on the upper right corner. Similarly, when utilizing the GFF: Subtraction strategy the $3^{rd}$ retrieved description shows the appearance of a single house in the lower left corner, misrepresenting both the quantity and the spatial location of the change. Although spatial references are often omitted in the sentences retrieved using TFF, the number of houses is correctly described, contributing to a better semantic match overall. As shown in Fig. \ref{Fig:I-T LEVIRCC_big} b), the EF baseline method retrieves the highest number of textual descriptions that so not align well with the query bitemporal image. Only one of the retrieved sentences correctly depict the appearance of the two buildings, while the others broadly describe changes related to the forest without specifying the constructed structures. These qualitative results further confirm the capability of the TFF strategy in modeling the semantic content of bitemporal images associated with change.

\subsection{Performance Variation between LEVIR CC and Dubai CCD}
A comparative analysis of the retrieval results in Tables~\ref{tab:dubai-full}- ~\ref{tab:dubai-nochange} and ~\ref{tab:LEVERCC_full}-~\ref{tab:LEVERCC_nochange} reveals notable differences in method performance across the Dubai CCD and LEVIR CC datasets, primarily due to variations in spatial resolution, annotation consistency, and sensor modality. Although TFF consistently outperforms EF and GFF variants on both datasets, absolute scores and performance gaps vary, especially for no-change queries. In the LEVIR CC dataset the bitemporal images associated to no-change are described by only 5 distinct sentences. This characteristic of the data set results in higher retrieval performance (compared to DUBAI CCD) and task asymmetry, simplifying $T \rightarrow I$ retrieval but increasing ambiguity in $I \rightarrow T$. In contrast, Dubai CCD consists of a much smaller number of image pairs and exhibits greater variability in the no-change descriptions. Combined with its lower-resolution multispectral imagery, this makes accurate retrieval more challenging for both tasks. We observe a divergent behavior between the two variants of the GFF. On LEVIR-CC, GFF: Concatenation performs better than GFF: Subtraction in the no-change subset, while the opposite trend is observed on Dubai CCD. This contrast can be attributed to the difference in the spatial resolution and modality of the images. LEVIR-CC’s high-resolution RGB images offer rich semantic details, making complementary features captured by Concatenation more effective. In contrast, Dubai CCD's lower-resolution multispectral images benefit more from the difference modeling introduced by Subtraction, which helps isolate subtle spatial and spectral variations. In the change query set, retrieval becomes more challenging on both datasets due to increased semantic complexity in the captions. As a result, the scores are generally lower for all methods. Nevertheless, with the TFF our method maintains the best overall performance, indicating its ability to model fine-grained temporal interactions through the cross-attention mechanism on both the dataset. These results highlight the importance of designing fusion strategies that generalize well across datasets, as well as the development of diverse, well-annotated benchmarks to better evaluate cross-modal text-ITSR in RS.

\subsection{Complexity Analysis}
This subsection presents the computational complexity of the EF baseline and our method with GFF variants and TFF. We report floating-point operations (FLOPs) per forward pass (inference) and corresponding total training time on the LEVIR-CC dataset(see Table~\ref{tab:complexity_analysis}). Among the compared approaches, the EF baseline is the most efficient, requiring 22.77B FLOPs per forward pass. Our method with the GFF variants (Concatenation and Subtraction) approximately doubles the computational cost to 45.08B FLOPs. Our method with the TFF is the most computationally intensive, with 54.36B FLOPs, roughly $20\%$ more than GFF variants. Despite the increased complexity and training time, with the TFF our method outperforms the EF method and the GFF variants across all evaluation metrics and query types (see Tables~\ref{tab:LEVERCC_full}-~\ref{tab:LEVERCC_nochange}), justifying the additional cost. We observe similar trends on the Dubai CCD dataset. However, we omit detailed results due to space constraints.

\begin{table}[ht]
\centering
\caption{Model complexity and training duration on the LEVIR-CC dataset. FLOPs are measured per forward pass (inference).}
\begin{tabular}{@{}lcc@{}}
\toprule
\textbf{Method} & \textbf{Total FLOPs (B)} & \textbf{Training Time (min)} \\
\midrule
EF Baseline & \num{22.77} & 18 \\
GFF: Concatenation & \num{45.08} & 31 \\
GFF: Subtract & \num{45.08} & 27 \\
TFF & \num{54.35} & 32 \\
\bottomrule
\end{tabular}
\label{tab:complexity_analysis}
\end{table}

\section{Conclusion}\label{conclusion}
In this paper, as a first time in RS we have introduced the text-ITSR task which aims to retrieve image time series using text sentences as queries and vice versa. In particular, we have presented a self-supervised text-ITSR method focusing on pairs of images (i.e., bitemporal images). Our method is composed of two  main components: 1) modality-specific encoders to extract discriminative features from bitemporal images and text sentences and 2) modality-specific projection heads to align textual and image features in a common embedding space using contrastive learning. To model temporal information at the feature level, we have used two fusion strategies: i) the global feature fusion (GFF) strategy, which combines global feature using two simple yet effective operators that are feature concatenation or feature subtraction; and ii) transformer-based feature fusion (TFF) strategy that leverages the transformer architecture for fine-grained information. In the experiments, we have evaluated our method by using two datasets under two tasks: i) RS image time series retrieval with a text sentence as query; and ii) sentence retrieval with a RS image time series as query. The experimental results show the effectiveness of our text-ITSR method in accurately retrieving bitemporal images given a query text sentence and vice versa. Among the fusion strategies, TFF is the most effective strategy for modeling temporal information within bitemporal images, leading to more accurate retrieval results across modalities compared to GFF strategies. This is more visible in the cases in which the queries (text sentences or bitemporal images) represent changes. This improved performance is due to the specific design of the TFF strategy to capture the changes within the bitemporal images. In addition, we included a comparison with a simple Early Fusion (EF) baseline method, where bitemporal images are stacked at the input level before feature extraction. While computationally more efficient, the EF achieves noticeably lower retrieval accuracy than our method, especially on change-related queries. This result proves the importance of explicitly modeling temporal relationships for an accurate text-ITSR. 

We would like to note that although in this work we have focused on text-ITSR problems, our architecture and training strategy have the potential to serve as an image time series-language foundation model when pre-trained in larger datasets. Through fine-tuning applied by using annotated image time series, the pre-trained modality-specific encoders could support various downstream tasks related to the analysis of image time series (e.g., change captioning, change related questioning answering). As a future development of our work, we plan to test our method in the context of retrieving long-term changes.

\ifCLASSOPTIONcaptionsoff
  \newpage
\fi

\bibliography{references}
\bibliographystyle{ieeetr}
\vfill

\end{document}